\pdfoutput=1

\documentclass[11pt]{article}

\usepackage{acl}

\usepackage{times}
\usepackage{latexsym}
\usepackage{graphicx}
\usepackage{ragged2e}
\usepackage{float}
\usepackage{longtable}
\usepackage{graphics}
\usepackage{multirow}
\usepackage{caption}
\usepackage{subcaption}

\usepackage[T1]{fontenc}

\usepackage[utf8]{inputenc}

\usepackage{microtype}

\title{Very Low Resource Sentence Alignment: Luhya and Swahili}

\author{Everlyn Asiko Chimoto \\
  University of Cape Town, South Africa \\
  African Institute for Mathematical\\ Sciences \\
  \texttt{everlyn@aims.ac.za} \\\And
  Bruce A. Bassett \\
  University of Cape Town, South Africa \\
  African Institute for Mathematical\\ Sciences, South Africa \\
  South African Astronomical Observatory\\
  \texttt{bruce.a.bassett@gmail.com } \\}

\begin{document}
\maketitle
\begin{abstract}
Language-agnostic sentence embeddings generated by pre-trained models such as LASER and LaBSE are attractive options for mining large datasets to produce parallel corpora for low-resource machine translation. We test LASER and LaBSE in extracting bitext for two related low-resource African languages: Luhya and Swahili. For this work, we created a new parallel set of nearly 8000 Luhya-English sentences which allows a new zero-shot test of LASER and LaBSE. We find that LaBSE significantly outperforms LASER on both languages. Both LASER and LaBSE however perform poorly at zero-shot alignment on Luhya, achieving just $1.5\%$ and $22.0\%$ successful alignments respectively (P@1 score). We fine-tune the embeddings on a small set of parallel Luhya sentences and show significant gains, improving the LaBSE alignment accuracy to $53.3 \%$. Further, restricting the dataset to sentence embedding pairs with cosine similarity above 0.7 yielded alignments with over $85\%$ accuracy.  
\end{abstract}

\section{Introduction}

Sentence alignment is the creation of parallel corpora from monolingual data~\citep{gale-church-1993-program,kay-roscheisen-1993-text}. This alignment can be done  manually and/or automatically. Manual alignment is laborious and costly hence there has been a lot of work on automatic sentence alignment~\citep{steingrimsson-etal-2021-effective,schwenk-2018-filtering,guo-etal-2018-effective}. Tasks such as Building Using Comparable Corpora (BUCC) focus on building parallel corpora using neural methods~\citep{zweigenbaum-etal-2017-overview,zweigenbaum:hal-01898360}. Essentially, sentences are aligned to the corresponding translation in another language using language agnostic sentence embeddings with the idea that sentences that are translations of each other will be close in the vector space~\citep{huang-etal-2015-translation,10.1145/2699927}. These sentence embeddings are generated using pre-trained models such as Language Agnostic Sentence Representation (LASER) and Language Agnostic BERT Sentence Embeddings (LaBSE)~\citep{schwenk-douze-2017-learning,artetxe-schwenk-2019-massively,feng-etal-2022-language}. LASER and LaBSE have been used to effectively mine bitext from comparable corpora.

As these pre-trained models are effective in mining bitext, we investigate how they would perform on an unseen low-resource language: Luhya, Marama dialect, as well as Swahili. Our main contributions are:
\begin{enumerate}
    \item We created a Luhya-English parallel corpus of nearly 8000 aligned Luhya\footnote{Also sometimes written as Luhyia.} and English sentences.
    \item An empirical evaluation of LASER and LaBSE on Luhya and Swahili datasets.
    \item Fine-tuning Luhya embeddings to improve bitext mining for this unseen language to explore the value of small amounts of parallel sentences for improving zero-shot performance. 
\end{enumerate}

\section{Multilingual Sentence Embeddings}

In this section, we review LASER and LaBSE.

\subsection{LASER}

\begin{figure*}[!ht]
\centering
		\includegraphics[height=0.23\textheight,width=0.98\textwidth]{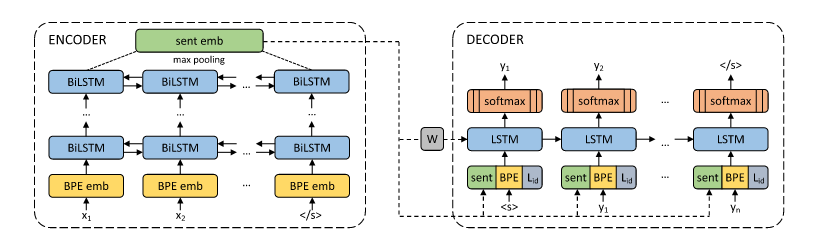}
\caption{The LASER architecture as proposed by~\citet{artetxe-schwenk-2019-massively} consisting of a single encoder and a single decoder. The encoder has 1-5 stacked BiLSTM layers followed by a max pooling. The decoder is an auxiliary component.}
		\label{fig:AL}
\end{figure*}

Language Agnostic SEntence Representation (LASER) is a framework used to obtain multilingual sentence embeddings~\citep{schwenk-douze-2017-learning}. It borrows from neural machine translation by utilizing encoders and decoders to generate the sentence embeddings which are of a fixed size in this case 1024~\citep{kalchbrenner-blunsom-2013-recurrent,DBLP:journals/corr/SutskeverVL14,cho-etal-2014-learning}.

The encoder-decoder architecture is shown in Figure \ref{fig:AL}. The encoder consists of 1-5 stacked BiLSTM layers each of dimension size 512 ~\citet{artetxe-schwenk-2019-massively}. The output of the encoder is max pooled to get the sentence embeddings. On the other hand, the decoder is an auxiliary component that consists of an LSTM layer of dimension 2048. LASER was trained by feeding 93 input languages to the system with a joint Byte Pair Encoding (BPE) with 50k merge operations. While the input to the encoder is just the BPE embedding, the input to the decoder consists of the sentence embedding generated by the encoder, the BPE embedding of the translation as well as the language ID. The encoder does not include the language ID since the goal is to  allow the model to learn language-independent representations.

At the time of its release, LASER achieved state-of-the-art results in mining bitext in the BUCC task dataset~\citep{zweigenbaum-etal-2017-overview,zweigenbaum:hal-01898360} for all language pairs except Chinese-English.

\subsection{LaBSE}
\begin{figure}[h!]
	\centering
    \resizebox{\columnwidth}{!}{
		\includegraphics[height=0.2\textheight,width=0.35\textwidth]{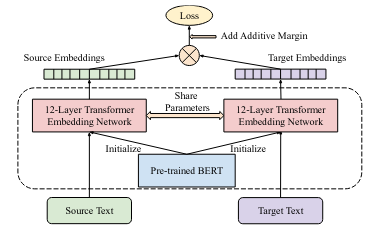}
		}
		\caption{The LaBSE architecture as proposed by ~\citet{feng-etal-2022-language} which uses a dual encoder, one taking in the source sentence while the other taking in the target sentence. The encoders are initialized using pre-trained BERT weights and the architecture is trained using the translation ranking loss with additive margin.}
		\label{fig:labse}
\end{figure}

The Language Agnostic BERT Sentence Embeddings (LaBSE) framework is a cross-lingual approach that utilises a pre-trained BERT model to generate sentence embeddings~\citep{feng-etal-2022-language,DBLP:journals/corr/abs-1902-08564}.
The LaBSE model consists of 12-layer transformer dual encoders which share parameters~\citep{guo-etal-2018-effective}. These encoders are initialized using pre-trained BERT weights~\citep{devlin-etal-2019-bert}. Each encoder is fed source and target text respectively and embeddings are trained by minimizing the translation ranking loss with additive margin softmax~\citep{Yang2019ImprovingMS}; see Figure \ref{fig:labse} for further details. Each output embedding vector has dimension 768. The LaBSE model was trained on 109 languages and achieved state-of-the-art performance with bitext mining as shown in \citet{feng-etal-2022-language,Heffernan2022BitextMU}.

\section{Languages}

In this section, we provide a brief description of the low-resource languages experimented on: Luhya (Marama dialect) and Swahili. The Marama dialect of the Luhya language, also known as Olumarama, is spoken in Western Kenya in Kakamega and Vihiga region with about 43,000 speakers~\cite{ethnologue}. The language status of Marama is educational as it is used vigorous both verbally and in broadcast media~\cite{ethnologue}. On the other hand, Swahili is spoken in East and Central Africa including countries such as Kenya, Tanzania, DRC, parts of Uganda \& Rwanda and has approximately 100 million speakers~\cite{ethnologue}. Its status is national as several countries use it as their national language~\cite{ethnologue}. These two languages are both Bantu languages from the Niger-Congo language family~\citep{LanguageFamilies}. Being from the same language family, they have the same word order structure, namely sentences follow a Subject-Verb-Object (SVO) ordering. They are also both agglutinative. Since both LASER and LaBSE were trained on Swahili, Luhya makes a very interesting zero-shot example to see how much information is transferred from the raw embeddings.

\section{Related work}

With the proliferation of neural embedding techniques, there have been various efforts to align sentences in various low-resource languages. \citet{thompson-koehn-2019-vecalign} introduce VecAlign which aligns sentences using LASER sentence embeddings~\citep{artetxe-schwenk-2019-massively} similarity as well as dynamic programming approximation. They experiment on low-resource language pairs namely: Sinhala-English and Nepali-English. They show that the sentences aligned using this method achieve improvement in machine translation models downstream. 

On the other hand, \citet{tien-etal-2021-kc4align} proposed KC4Align that utilises a multilingual translation system to generate embeddings. The similarity in these embeddings is used to perform paragraph alignment. Sentences are aligned where the sentences appear in the paragraph alignment using similarity scores and sentence length ratio. This method was tested on the Vietnamese-Laos language pair. Focusing on African low-resource language,  \citet{schwenk-etal-2021-wikimatrix} extracts parallel sentences from Wikipedia using multilingual sentences embeddings for Swahili among other languages. 

Regarding Luhya, there has been work on building Luhya datasets. \citet{steimel-2018-part} work focuses on parts of speech tagging for the Wanga Luhya dialect whereas \citet{DVN/NOAT0W_2022} focuses on producing parallel datasets for several Luhya dialects to English with the help of human translators. This is in contrast to our work which analyses automatic sentence alignment for Luhya. 

\section{Methodology}

\subsection{LASER and LaBSE evaluation}

LASER and LaBSE were utilised to generate our raw embeddings. Following the embedding generation, we compute the cosine similarity and the Euclidean distance (L2) from each vector in the English embedding set and all the Luhya/Swahili embeddings. Sentences are aligned to the most similar or closest sentence. We took both the Top-1 and Top-3 best alignments to test the performance of the pre-trained models. 

We use accuracy as our key metric. An important note is that we evaluate alignment performance by demanding exact matching of sentence indices on both sides. This means that if a sentence appears more than once in one of the languages, and the alignment chooses the ``wrong" index, despite the sentence being identical to the ``correct" sentence, then this is classified as a fail. This will apply primarily to the Top-1 results. As a consequence, our accuracy estimates should be taken as a lower bound on the true alignment performance.   

\subsection{Fine-tuning the embeddings}

\begin{figure}[ht!]
	\centering
    \resizebox{\columnwidth}{!}{
		\includegraphics[height=0.2\textheight]{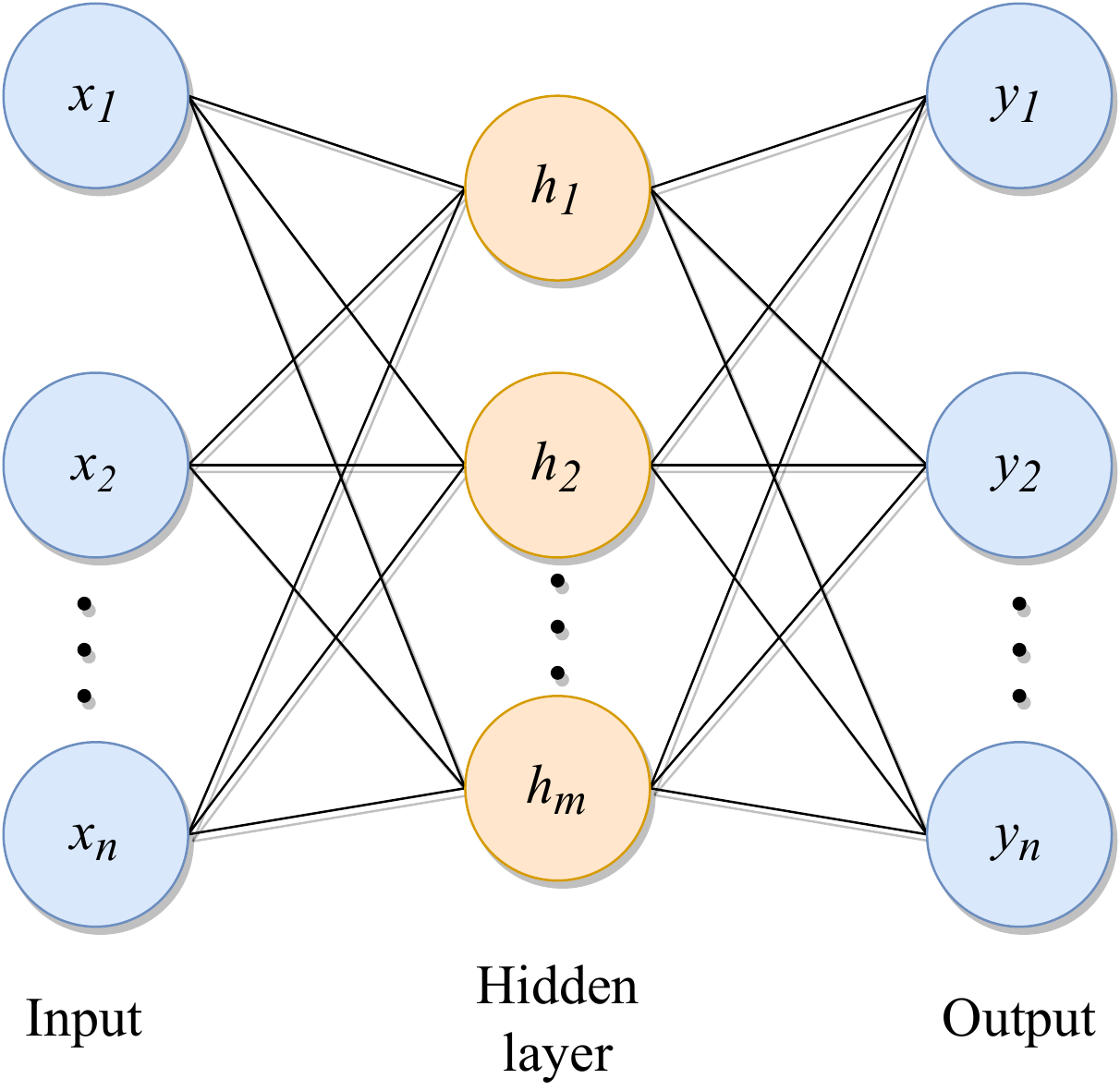}
		}
		\caption{Fine-tuning architecture with one bottleneck layer that can vary in size. Hidden sizes of 32, 64, 96, 128 and 256 were tested. The network takes in the generated embeddings from LaBSE and outputs new vectors that maximize the similarity between the new Luhya embedding and English embedding.}
		\label{fig:fine-tune-network}
\end{figure}

To fine-tune our Luhya embeddings, we added a fully-connected network with a single hidden layer to help learn new weights where the cosine similarity between the new embedding and the English embedding would be greatest. We defined the loss as:

$$ \textrm{Loss}(\mathbf{x},\mathbf{y}) = 1 - S_C(\mathbf{\tilde{x}},  \mathbf{y})$$
where $\mathbf{x},\mathbf{y}$ are the raw Luhya and English embeddings, $S_C$ is the cosine similarity and $\mathbf{\tilde{x}}$ are the fine-tuned Luhya embeddings which depend on $\mathbf{w}_{1,2}$, the vectors of new weights introduced by the fine-tuning architecture~\citep{Pal2017CopernicanL}. $\mathbf{w}_{1}$ representing weights to the hidden layer and $\mathbf{w}_{2}$ representing weights to the output layer. The bottleneck layer size is a hyper-parameter that we vary to explore its impact on performance; see Figure \ref{fig:fine-tune-network}.

\section{Experiments}

\subsection{Datasets and Alignment}

We experiment on both Luhya-English  and  Swahili-English sentence alignment.  The Luhya-English parallel set was created by aligning sentences from the New Testament Bible translations. This dataset consists of 7952 parallel sentences in the Marama dialect of Luhya. This bitext creation was achieved by cleaning, curating and aligning the New Testament of the Bible in Luhya and English. After alignment, the dataset was assessed by three independent Native speakers to assess the quality of alignment. This dataset is the only known digital parallel corpus in Luhya-English for the Marama dialect. See examples of the aligned sentences on Table \ref{tab:example}.

\begin{table*}[ht]
\centering
\begin{tabular}{|p{0.46\linewidth} | p{0.46\linewidth}|}
\hline
\multicolumn{1}{|l|}{\textbf{Luhya}}                                                                                                                 & \multicolumn{1}{l|}{\textbf{English}}                                                                            \\ \hline
Nebutswa omukholi , womumukunda oyo namukalusia ari , ‘ Omwami , lekha , kubekhwoho omuyika kuno khandi , nasi ndalakwachila , nekurakhwo imbolela. & But he answered and said to him , ‘ Sir , let it alone this year also , until I dig around it and fertilize it. \\ \hline
Ne , nali emakombe nanyasibungwa muno , yahenga ikulu ne , nalola Aburahamu nende Lazaro nibali halala ehale.                                    & And being in torments in Hades , he lifted up his eyes and saw Abraham afar off , and Lazarus in his bosom.     \\ \hline
Saulo namenya ninabo nayaala muliira lia Yesu , mu Yerusalemu obularia likhuwa liosi liosi tawe.                                                  & So he was with them at Jerusalem , coming in and going out. \\ \hline                                                  
\end{tabular}
\caption{Sample aligned sentences from our bible dataset.}
\label{tab:example}
\end{table*}

The Swahili-English dataset was sourced from the SAWA dataset which contains approximately $89k$ parallel sentences from various domains \citep{de-pauw-etal-2009-sawa,Pauw2011ExploringTS}. We sampled 10k parallel sentences from the Bible. This sampling allows comparison of automatic alignment specifically in the religious domain. Data cleaning involved getting rid of characters from different text encodings, removing both extra white spaces and verse numbers from all the datasets.

\subsection{Results and Discussion}

\begin{table*}[ht]
\centering
\begin{tabular}{cl|ll|ll}
\multicolumn{1}{l}{}              &                            & \multicolumn{2}{c}{\textbf{LASER}} & \multicolumn{2}{|c}{\textbf{LaBSE}} \\
\multicolumn{1}{l}{}              &                            & \textit{Top-1}   & \textit{Top-3}  & \textit{Top-1}   & \textit{Top-3}  \\ \hline \hline
\multirow{2}{*}{\textbf{Luhya-English}}   & \textrm{Cos. Sim.} & 0.02           & 0.02          & 0.22           & 0.32          \\
                                  & \textrm{L2}               & 0.00           & 0.01          & 0.22           & 0.32          \\ \hline
\multirow{2}{*}{\textbf{Swahili-English}} & \textrm{Cos. Sim.} & 0.50           & 0.55          & 0.97           & 1.00          \\
                                  & \textrm{L2}               & 0.45          & 0.55         & 0.97           & 1.00   \\ \hline       
\end{tabular}
\caption{Alignment accuracy for Luhya and Swahili using the raw LASER and LaBSE embeddings (no fine-tuning). The Top-1 (Top-3) columns represent the accuracy of correctly aligned sentences based on the sentences with the top 1 (3) most similar embeddings based either on cosine similarity ("Cos. Sim") or Euclidean distance ("L2"). LaBSE performs better than LASER on both languages,  correctly aligning 22.0\% of Luhya-English sentences and 97.1\% of Swahili-English sentences. LASER performs poorly when aligning Luhya-English with only 1.5\% being aligned correctly.  }
\label{tab:result}
\end{table*}

We utilize LASER and LaBSE to test out zero-shot bitext mining on Luhya-English dataset. Luhya is not included in the initial training of these models. We take both cosine similarity and Euclidian distance of the English embedding to the Luhya embedding. The results can be seen on table \ref{tab:result} where we can see we see LaBSE outperforming LASER on Luhya-English alignment by matching up to 22\% of the sentences correctly whereas LASER only matched 0.02\%. We also note that increasing the number of sentences that are match from 1 to 3 does not increase the performance of LASER in bitext mining while LaBSE performance increases by ~10\%. The top-3 result means we consider accurate alignment if the correct alignment was among the top 3 matched sentences. 

The performance of LaBSE shows that there are great gains achieved by utilising a pre-trained model in the sentence embedding model. LaBSE model utilises BERT in its training offering cross-lingual benefit that results in up to 22\% accurate alignment on a language it has not seen before. LASER on the other hand was trained from scratch and does not provide great results in aligning Luhya, an unseen language.

Considering the performance on Swahili, we see that prior knowledge of a language greatly helps in performance. Swahili performs better than Luhya in the alignment with LaBSE embeddings resulting in near perfect alignment (See table \ref{tab:result}). The performance of LASER with Swahili does not correspond with the results by \citet{artetxe-schwenk-2019-massively} where F1 scores of above 90\% were recorded. In our case, the F1 score is equivalent to the accuracy as the number of extracted parallel sentences is equivalent to the number of gold standard alignments. Contrary to the LASER results, the results of LaBSE outperforming LASER corresponds with the results from~\citet{feng-etal-2022-language,Heffernan2022BitextMU}. We also observed that cosine similarity performs marginally better than Euclidean distance on average and hence is used for our fine-tuning experiments.

\subsection{Fine-tuning LaBSE Luhya embeddings}

Owing to the good performance of 22\% on zero-shot alignment, we fine-tune the LaBSE Luhya embedding to evaluate the extent one needs to go to see improvements. Initially, we added one additional layer without the bottleneck and trained this network with 50\% of the Luhya dataset while testing on the other 50\%. This achieved an accuracy of 40.22\%. However, we did not pursue this network further as the number of trained parameters was too large to offer value for a small data as the Luhya dataset. We added a bottleneck layer whose input was the 768-sized embedding and the output layer was of size 768. Our experimental setup aimed to investigate what amount of correctly aligned sentences are needed to fine-tune the embeddings and see improvement as well as what is the optimal hidden size to achieve improvement in alignment. We split the Luhya-English dataset into 5-folds. At each iteration of training, one fold of 1591 parallel sentences was used for testing while the other folds were used for training. To investigate how many sentences are required to see improvements, we used 10\% of the training set to fine-tune the embeddings and evaluated the model with the test set and continuously added 10\% until the whole training set was used up. The training set consisted of 6361 parallel sentences. 

\begin{figure}[ht!]
	\centering
    \resizebox{\columnwidth}{!}{
		\includegraphics[height=0.2\textheight,width=0.35\textwidth]{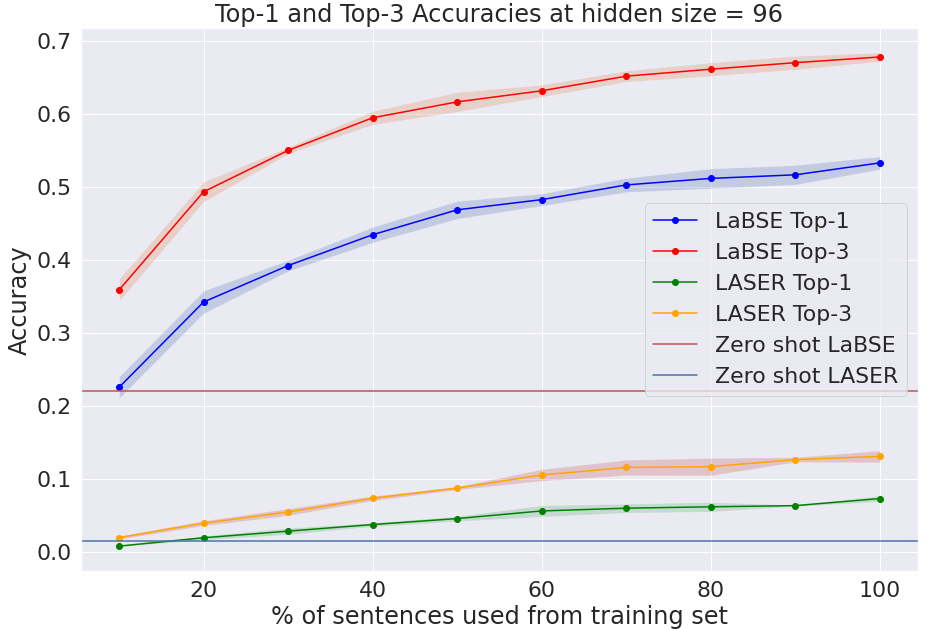}
		}
		\caption{LaBSE outperforms LASER in both the Top-1 and Top-3 results from fine-tuning the respective embeddings (with a hidden layer  dimension of 96) for Luhya. Results are shown as a function of the percentage of the full training set of 6361 sentences used for training. For LaBSE the Top-1 (Top-3) accuracy reaches 53\%  (68\%). Error bars are standard deviations estimated from 5-fold cross-validation.}
		\label{fig:finresults222}
\end{figure}

\begin{figure}[ht!]
	\centering
    \resizebox{\columnwidth}{!}{
		\includegraphics[height=0.2\textheight,width=0.35\textwidth]{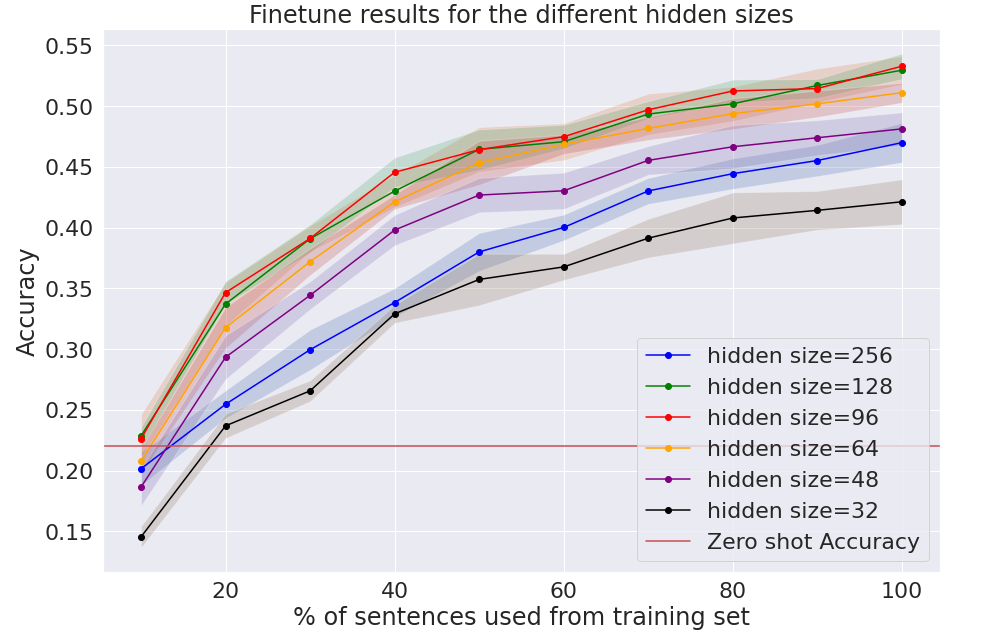}
		}
		\caption{Top-1 results from fine-tuning the Luhya embeddings from LaBSE using different hidden layer sizes, as a function of the percentage of the full training set of 6361 sentences, together with the zero-shot result of $22.0\%$. Results are the mean of 5-fold cross validation with error bands given by $\pm 1\sigma$, where $\sigma$ is the standard deviation of the 5 folds. The best hidden sizes are for hidden layers of dimensionality 96 and 128.}
		\label{fig:finresults}
\end{figure}

\begin{table}
\centering
\resizebox{\columnwidth}{!}{
\begin{tabular}{l|l|lll}
\textbf{Training Size}    & \textbf{Threshold} & \textbf{Precision} & \textbf{Recall} & \textbf{F1 Score} \\ \hline \hline
\multirow{3}{*}{At 10\%}  & -0.2               & 0.22               & 0.22            & 0.22              \\
                          & 0.54               & 0.29               & 0.18            & 0.22              \\
                          & 0.68               & 0.54               & 0.03            & 0.06              \\ \hline
\multirow{3}{*}{At 20\%}  & -0.2               & 0.34               & 0.34            & 0.34              \\
                          & 0.5                & 0.38               & 0.31            & 0.34              \\
                          & 0.69               & 0.72               & 0.05            & 0.09              \\ \hline
\multirow{3}{*}{At 40\%}  & -0.2               & 0.44               & 0.44            & 0.44              \\
                          & 0.52               & 0.49               & 0.41            & 0.45              \\
                          & 0.72               & 0.83               & 0.06            & 0.11              \\ \hline
\multirow{3}{*}{At 60\%}  & -0.2               & 0.48               & 0.48            & 0.48              \\
                          & 0.53               & 0.54               & 0.43            & 0.48              \\
                          & 0.73               & 0.88               & 0.06            & 0.11              \\ \hline
\multirow{3}{*}{At 80\%}  & -0.2               & 0.51               & 0.51            & 0.51              \\
                          & 0.5                & 0.55               & 0.49            & 0.52              \\
                          & 0.75               & 0.9                & 0.05            & 0.09              \\ \hline
\multirow{3}{*}{At 100\%} & -0.2               & 0.53               & 0.53            & 0.53              \\
                          & 0.5                & 0.56               & 0.51            & 0.53              \\
                          & 0.75               & 0.92               & 0.05            & 0.09         \\ \hline    
\end{tabular}}
\caption{Precision, Recall and F1 score as the cosine similarity score threshold is increased with different training sizes. -0.2 threshold represents not setting a threshold at all, this results in all sentences being aligned however the precision is very low as some sentences are wrongly aligned. As the threshold increases fewer sentences are considered aligned, however there are more accurate alignment. Hence, the precision increases while the recall and F1 score decrease.  }
\label{tab:precision}
\end{table}

Figure \ref{fig:finresults} shows the results in which we see that regardless on the hidden size only 20\% of the training data set, about 1272 sentence, is sufficient to result in the improvement of the alignment. Also, looking at the different hidden sizes 128 and 96 hidden sizes were the best with no distinct difference between the two. As much as the hidden size of 128 is comparable with 96, hidden size of 96 works with fewer parameters thus more efficient. Increasing the hidden size beyond 128 degraded the performance as evidence by the performance of hidden size 256. Also, smaller hidden sizes did not offer much gain.

As we note good results with a hidden size of 96, we perform both the top-1 and the top-3 evaluation for both LASER and LaBSE Luhya embedding fine-tuning. Figure \ref{fig:finresults222} shows that fine-tuning with hidden size 96 results in an accuracy of up to 68\%. These results show that with little effort, LaBSE embeddings can be used to effectively mine bitext of languages not seen during training. This is practical for various very low-resource languages.


\begin{table*}[ht]
\centering
\begin{tabular}{l|llllllllll}
               & \textbf{10\%} & \textbf{20\%} & \textbf{30\%} & \textbf{40\%} & \textbf{50\%} & \textbf{60\%} & \textbf{70\%} & \textbf{80\%} & \textbf{90\%} & \textbf{100\%} \\ \hline
\textbf{Top-1} & 0.22         & 0.34         & 0.39         & 0.43         & 0.47         & 0.48         & 0.50         & 0.51         & 0.52         & {\bf 0.53}          \\
\textbf{Top-3} & 0.36         & 0.49         & 0.55         & 0.59         & 0.62         & 0.63         & 0.65         & 0.66         & 0.67         & {\bf 0.68}         
\end{tabular}
\caption{Fine-tuning results with a hidden layer size of 96 for the LaBSE embeddings. With 100\% of the training data, (6362 sentences), the Top-1 accuracy is 53.2\%, while the Top-3 accuracy is 67.8\%.}
\label{tab:96}
\end{table*}

\begin{figure}[ht!]
	\centering
    \resizebox{\columnwidth}{!}{
		\includegraphics[height=0.2\textheight,width=0.35\textwidth]{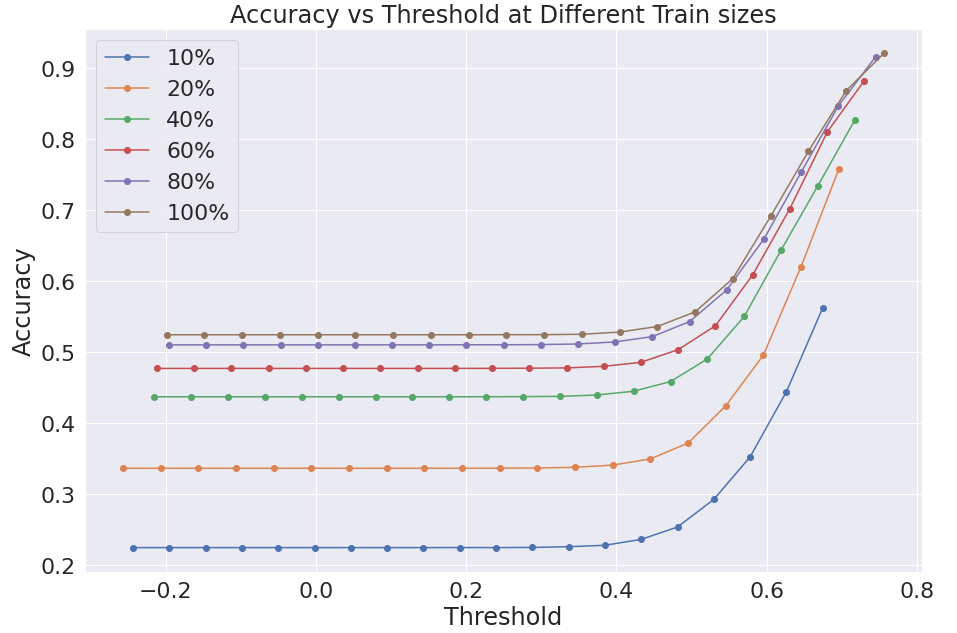}
		}
		\caption{Top-1 accuracies as a function of cosine similarity threshold and training dataset percentage. We see that accuracy correlates with similarity, allowing the curation of high-accuracy datasets suitable for use in machine translation.}
		\label{fig:finresults2}
\end{figure}

To assess the accuracy of aligned sentences after fine-tuning, we analyse the accuracy against different similarity score thresholds along with different training set size. Figure \ref{fig:finresults2} shows that fine-tuning with 40\% of the training dataset gets an accuracy of above 80\% meaning with ~2500 sentences one can get >80\% accuracy in alignment past 0.65 threshold. Table \ref{tab:precision} shows that setting a high similarity threshold results in higher precision but we see a significant drop in recall. Using the full training set, we see that without setting a threshold, the precision and recall are 53\%. However, setting the threshold to 0.75 improves the precision to 92\% while the recall drops to just 5\%. Setting a higher threshold results in more accurate alignments but selects fewer sentence pairs. Choosing the optimal threshold to balance precision and recall will depend on the task being considered. 

\section*{Conclusion}

Low-resource languages lack digitized parallel data needed for developing machine translation models. Sentence embedding models offer a potential way to create parallel data cheaply for these low and very low-resource languages but themselves need parallel data for their training. 

In this work we present a new dataset consisting of 7952 sentences translating the Luhya (Marama) dialect into English. We use this data, together with a similar Swahili dataset, to explore transfer learning and fine-tuning based on raw embeddings from the LASER and LaBSE algorithms for alignment on these languages. 

We show that LaBSE significantly outperforms LASER on both Swahili and Luhya but that both struggle with zero-shot learning on Luhya, achieving alignment accuracies of $22.0\%$ and $1.5\%$ respectively. We also show that fine-tuning with as little as 1200 correctly aligned Luhya sentences can result in models with significantly improved sentence alignments. In addition, setting a minimum similarity score threshold results in datasets with much more accurate alignments, useful for curating high-quality parallel corpora for machine translations. However, this comes at the cost of significantly reducing the number of aligned sentences.  We leave it to future work to investigate active learning for the choice of sentences for fine-tuning that will result in the greatest gains in performance. 

\section*{Acknowledgements}

The authors gratefully acknowledge the contributions of Joy Nyende, Roger Nyende and Stephen Wakhu for their help in curating the Luhya Bible dataset. We thank Emmanuel Dufourq and the anonymous reviewers for comments and contributions to the draft. This publication was made possible by a grant from Carnegie Corporation of New York (provided through the African Institute for Mathematical Sciences). The statements made and views expressed are solely the responsibility of the authors.

\bibliography{anthology,custom}




\end{document}